\documentclass{article}

\usepackage{microtype}
\usepackage{graphicx}
\usepackage{subfigure}
\usepackage{booktabs} % for professional tables
\usepackage{float}

\usepackage{hyperref}

\usepackage[accepted]{icml2019}

\begin{document}
\twocolumn[
\icmltitle{Efficient Machine Learning for Large-Scale Urban Land-Use Forecasting in Sub-Saharan Africa}

% It is OKAY to include author information, even for blind
% submissions: the style file will automatically remove it for you
% unless you've provided the [accepted] option to the icml2019
% package.

% List of affiliations: The first argument should be a (short)
% identifier you will use later to specify author affiliations
% Academic affiliations should list Department, University, City, Region, Country
% Industry affiliations should list Company, City, Region, Country

% You can specify symbols, otherwise they are numbered in order.
% Ideally, you should not use this facility. Affiliations will be numbered
% in order of appearance and this is the preferred way.
\icmlsetsymbol{equal}{*}

\begin{icmlauthorlist}
\icmlauthor{Daniel Omeiza}{to}
\end{icmlauthorlist}

\icmlaffiliation{to}{Carnegie Mellon University}
\icmlcorrespondingauthor{Daniel Omeiza}{daniomeiza@gmail.com}
% You may provide any keywords that you
% find helpful for describing your paper; these are used to populate
% the "keywords" metadata in the PDF but will not be shown in the document
\icmlkeywords{Machine Learning, ICML}

\vskip 0.3in
]
% this must go after the closing bracket ] following \twocolumn[ ...

% This command actually creates the footnote in the first column
% listing the affiliations and the copyright notice.
% The command takes one argument, which is text to display at the start of the footnote.
% The \icmlEqualContribution command is standard text for equal contribution.
% Remove it (just {}) if you do not need this facility.

\printAffiliationsAndNotice{}  % leave blank if no need to mention equal contribution
% \printAffiliationsAndNotice{\icmlEqualContribution} % otherwise use the standard text.

\begin{abstract}
Urbanization is a common phenomenon in developing countries and it poses serious challenges when not managed effectively. Lack of proper planning and management may  cause the encroachment of urban fabrics  into reserved or special regions which in turn can lead to an unsustainable increase in population. Ineffective  management and planning generally leads to depreciated standard of living, where physical hazards like traffic accidents and disease vector breeding become prevalent. In order to support urban planners and policy makers in effective planning and accurate decision making, we investigate urban land-use in sub-Saharan Africa. Land-use dynamics serves as a crucial parameter in current strategies and policies for natural resource management and monitoring. Focusing on Nairobi, we use an efficient deep learning approach with patch-based prediction to classify regions based on land-use from 2004 to 2018 on a quarterly basis. We estimate changes in land-use within this period, and using the Autoregressive Integrated Moving Average (ARIMA) model, our results forecast land-use for a given future date. Furthermore, we provide labelled land-use maps which will be helpful to urban planners.
\end{abstract}

\section{Introduction}
\label{submission}
There have been rapid infrastructural developments in major cities in sub-Saharan Africa (especially in Nairobi) in terms of significant increase in residential houses, factories, and office buildings. These developments have led to a large increase in population and, in turn, to overcrowding. Africa's population is projected to almost triple by the year 2050 \cite{donnelly}. This increase will occur primarily in urban areas so that by 2025 about 800 million people will live in urban communities \cite{donnelly}.

While urban development was generally believed to improve living lifestyle, and  reduce risk of disease vector breeding, rapid increase in population, if not properly managed, usually presents series of challenges. These challenges include reduced standard of living, poor housing, inadequate environmental sanitation facilities which increase vector breeding and human contact, and road traffic congestion which may lead to road accidents.

In order to inform the decisions of investors and policy makers in Nairobi on urbanization, we study urbanization in Nairobi by analyzing historical satellite imagery of certain regions in Nairobi and forecast future land-use in these regions. We provide labelled land-use maps that will be helpful in urban planning.

The approach involves using a deep Convolutional Neural Network (CNN) classifier to do patch-based land-use predictions in order to classify land-use into the following ten classes: commercial area, dense residential, medium residential, spatial residential, parking lot, freeway, round-about, meadow, chaparral, and open space(bareland, wetland). We cannot apply object-based classification, for instance detecting each type of plant as in \cite{daniel} to determine a vegetation class. This consumes more compute resources and takes more time.

We detect changes in each land-use class (in square kilometers) and show changes visually through image differencing. We also compute percentage change per land-use class for each successive (temporal) pair of images. Using Autoregressive Integrated Moving Average (ARIMA) \cite{arima}, we forecast land-use for future dates.

\section{Importance and Prior Work}
In this section, we briefly review previous approaches that used various deep learning techniques in land-use classification and we examine possible training datasets for land-use classification.

\subsection{Land-Use Classification}
Efforts have been made over the years in developing various methods for the task of remote sensing image scene classification. Image scene classification is concerned with categorizing scene images into a discrete sets of meaningful Land-Use and Land Cover (LULC) classes according to the image contents \cite{cheng}. This has the potential of helping natural hazards detection and disaster recovery processes \cite{martha, cheng2, stumpf}, LULC determination \cite{wu2}, geospatial object detection \cite{cheng3}, geographic image retrieval \cite{wang, liY}, vegetation mapping \cite{kim, li}, environment monitoring, and urban planning.

Albert \cite{albert} used computer vision techniques based on deep CNNs for the classification of satellite imagery in order to identify patterns in urban environments using large-scale satellite imagery data. Ground truth land-use class labels were sampled from the Urban Atlas land classification dataset acquired through open-source surveys. The urban atlas dataset comprises of 20 land-use classes from about 300 European cities.
This data was used to train and compare deep architectures with ResNet-50 \cite{he2} exhibiting the best performance. ResNet-50 is a residual learning framework which is substantially deeper than previous networks. Unfortunately, models trained with the Urban Atlas dataset are unlikely to generalize well to other part of the world since training samples were acquired from only European cities. Consequently, the training dataset must be carefully selected to ensure that it can be applied successfully in sub-sahara Africa.

\subsection {Training Dataset}
Data is required for training classification networks. We require an urban land-use dataset that is representative of the target geography, that has a sufficient geographical coverage, that has rich variation of image data, and that possesses high within class diversity and between class similarity.

\subsubsection{DeepSat Dataset}
The DeepSat \cite{basu} dataset was released in 2015 and it contains two benchmarks which include the Sat-4 data of 500,000 images over 4 land use classes (barren land, trees, grassland, other), and the Sat-6 data of 405,000 images over 6 land use classes (barren land, trees, grassland, roads, buildings, water bodies). All the samples have a size of \(28 \times 28\) pixels at a 1m per pixel spatial resolution with 4 channels (red, green, blue, and NIR - near infrared). CNN models trained on this dataset have perform well during classification. While useful as input for pre-training more complex models, (e.g., image segmentation), the dataset lack diverse land-use classes, hence, not suitable for  detailed land use analysis and comparison of urban environments across cities.

\subsubsection{NWPU-RESISC45 Dataset}
The North-West Polytechnic University  (NWPU-RESISC45) dataset \cite{cheng} consists of 31,500 remote sensing images divided into 45 scene classes. Each class is made up of 700 images with a size of \(256 \times 256\) pixels in the Red Green Blue (RGB) color space. The spatial resolution of the images varies from about 30 m to 0.2 m per pixel except for the images that belong to the island, lake, mountain, and snowpack class. The 31,500 images cover more than 100 countries and regions all over the world, including developing, transition, and highly developed economies. In contrast to other existing datasets such as UC Merced Land-Use dataset, WHU-RS19 dataset \cite{sheng}, SIRI-WHU dataset \cite{zhao3}, RSSCN7 dataset \cite{zou}, RSC11 dataset \cite{zhao2}, and DeepSat dataset \cite{basu}, the NWPU-RESISC45 dataset is large scale, it possesses rich image variation, and it has high within class diversity and between class similarity. This dataset therefore provides the best coverage of the required characteristics.

\subsection{Data Collection}
Nairobi is one the African countries with high urbanization rate. In fact, Nairobi was ranked among the top 20 most dynamic cities in the world in the 2018 JLL’s Global Real Estate Transparency Index, it was the only city in Africa considered within this position. This is why have chosen Kilimani region of Nairobi as our region of interest in this research. In order to forecast urban land-use, some data about the region of interest is required. To obtain land-use dataset of Kilimani, we reviewed 3 aerial imagery sources, which include LandSat8 satellite, Sentinel-2 satellite, and Google Earth Software. We chose Google Earth's software as our source of aerial images because it possesses free high resolutution images and it also has rich collection of aerial images from 2014 to 2019. We retrieved images of  Kilimani region of interest  from 2004 to 2018 on a quarterly basis. This amounted to 60 aerial images.

\subsection{Data Pre-Processing}
Several factors affect the quality of satellite imagery. Some of these factors include cloud cover in the area of interest at the time of capture, inconsistent gaps in image time-series dataset, and low hardware capacity. All these factors can cause acquisition of low quality images. Pre-processing dataset to remove clouds and interpolating dataset to fill gaps is important for deep learning tasks. We removed images totally covered by cloud, from our dataset and we interpolated to make up for missing dates.

\subsection{Unsuitability of Google Earth's Images for Direct Training}
The nature of the images from Google Earth's desktop software dismisses the idea of directly training a simple deep learning model on them. This is because they are of high spatial and pixel resolutions. Furthermore, they do not contain labels for training, thus, we examined 8 possible training datasets for land-use classification using the metrics in Section 2.2.2 for best fit. The North West Polytechnic University land-use (NWPU-RESISC45) dataset \cite{cheng}) was the best fit for the classification tasks at hand. We hypothesized that a model trained on the NWPU-RESISC45 dataset should be able to predict land-use on the Google Earth's dataset using our proposed patch-based prediction technique, because of the similarities in the land-use classes in both datasets.

\subsection{Suitability of NWPU-RESISC45 Dataset}
We discuss the suitability of the NWPU-RESISC45 dataset for training a model for predicting aerial images of Kilimani retrieved from Google Earth's software. The following points will be considered: dataset geographical distribution, dataset size, image variation, high within class diversity and between class similarity, and the dataset distribution.

\paragraph{Dataset Geographical Distribution}
The region under consideration is Kiliimani in Kenya. Kenya happens to be a developing country, therefore it is important that our dataset for training the classification network contains samples from developing region in order to get obtain a good classification accuracy. NWPU-RESISC45 dataset contains 31,500 images from more than 100 countries and regions all over the world, including developing, transition, and highly developed economFies. This makes the dataset representative.

\paragraph{Dataset Size}
A contributing factor to the good performance of deep learning algorithms is the presence of large-scale dataset. NWPU-RESISC45 dataset is 15 times larger than the well popular UC Merced land-use dataset, it has about the largest scale on the number of scene classes. Also, each image has a dimension of 256 x 256 pixels, this makes provision for more information that will enhance the learning process of a deep CNN network.

\paragraph{Image Variation}
Tolerance to image variations is an important and well desired property of any scene classification system, be it human or machine \cite{cheng}. Unfortunately, many of the existing dataset are not rich with varieties of samples for same class. In contrast, the NWPU-RESISC45 images were carefully selected under all kinds of weathers, seasons, illumination conditions, imaging conditions, and scales \cite{cheng}. Thus, for each scene class, the dataset possesses richer variations in translation, viewpoint, object pose and appearance, illumination, background, spatial resolution, and occlusion.

\paragraph{High within Class Diversity and between Class Similarity } due to low class diversity and low between class similarity in dataset, some top-performing methods built upon deep neural networks have achieved saturation of classification accuracy. In contrast, NWPU-RESISC45 dataset contains high within class diversity and between class similarity making it good for land-use classification tasks \cite{cheng}.

\paragraph{Data Distribution}
The 31,500 image dataset contains 45 classes with 700 images in each class.
We selected 10 classes out if the 45 classes based on the common important land-use classes obtainable in developing regions. The selection of only 10 classes also helped to avoid class overlaps. The selected 10 land-use classes are: commercial area, dense residential, medium residential, spatial residential, parking lot, freeway, road-junction, meadow, chaparral and open-space(dryland, wetland or vegetation). Thus, our dataset contains 7000 images \((700 \times 10)\).

\section{Patch-Based Prediction}
We propose a patch-based prediction technique for classification and segmentation. A patch in this case is a rectangular super-pixel from an image. For a given patch, we apply a classification network, we obtain the maximum probability class, and color the tile according to that class; this results into a segmentation map. Each patch has the pixel dimension of the training images (\(256 \times 256 \) pixel in our case). A patch is selected by convolving each image at a fixed stride. The stride is the number of pixels by which we shift at each instance to select a new patch from an image. The smoothness of the segmentation map depends on the value of the stride value used (the smaller the better). We state the patch-based prediction algorithm in Algorithm \ref{alg:example}. In convolving, we start from the pixel at co-ordinate \((0,0)\) and select a patch based on a specified dimension, predict the class of the selected patch, color the pixels (based on its class) in the corresponding location of this patch in a copy of the original image.
\begin{algorithm}[tb]
   \caption{Patch-Based Prediction.}
   \label{alg:example}
\begin{algorithmic}
   \STATE {\bfseries Input:} $test\_img$, $empty\_matrix$, $size$, $stride$
   \STATE Initialize $A \Leftarrow test\_image$
   \STATE Initialize $E \Leftarrow empty\_matrix$
   \FOR{$x=0$ {\bfseries to} $Width - size$ by $stride$}
   \FOR{$y=0$ {\bfseries to} $Height - size$ by $stride$}
    \STATE $patch \Leftarrow A[x$ {\bfseries to} $  x+size-1][y$ {\bfseries to} $y+size-1][0$ {\bfseries to} $2]$
    \STATE $preprocess\_and\_reshape(patch)$
    \STATE $class \Leftarrow predict(patch)$
    \STATE $color(E[x$ {\bfseries to}  $x+size-1][y$  {\bfseries to}  $y+size-1][0$ {\bfseries to}  $2], class\_color)$
   \ENDFOR
   \ENDFOR
\end{algorithmic}
\end{algorithm}

Aerial images retrived from Google Earth's software are of high pixel and spatial resolution. Performing semantic segmentation on high resolution images requires masked images as labels during training. Creating mask labels for land-use images is expensive and takes much time. In fact, such labelled dataset are not easily accessible and do not even exist for some developing countries. Also, model architectures for semantic segmentation may be complex and may therefore take more training time with high resolution images (\(3200 \times 4800\) pixels in our case). Patch-based prediction exhibits better efficiency in this case. We therefore train a simple deep CNN network with a regular land-use dataset (NWPU-RESISC45 dataset with \(256 \times 256 \) pixels images, and text labels rather than masked image labels) and perform patch-based prediction on the high resolution images after training, to obtain segmentation maps.

\section{Implementation of the Classification System }
In this section, we discuss the chosen performance benchmark, model architecture design and the training procedure.

\subsection{Benchmark Model}
In order to provide a performance benchmark for the intended classification model, we require a benchmark model performance which will inform us of the relative performance of our model. We use VGGNet-16 performance recorded in \cite{cheng} as a benchmark. Cheng et. al \cite{cheng} evaluated few pre-trained deep CNN models such as AlexNet, VGGNet-16, and GoogLeNet on the NWPU-RESISC45 dataset after fine-tuning the models. The authors used 20\% of the entire dataset to train these networks. It turns out that VGGNet-16 (one of the top performing image pre-trained classification model on imagenet) produced the highest overall accuracy of 90.36\%.   The overall accuracy is defined as the number of correctly classified samples, regardless of which class they belong to, divided by the total number of samples. We expect our model to reach or surpass this performance.

\subsection{Model Architecture}
We used two approaches: transfer learning with ResNet-50 and a convolutional neural network designed from scratch.
In the transfer learning approach, we performed fine-tuning by removing the top layer of ResNet-50 network (i.e the fully connected and the classification layer), then added GlobalAveragePooling layer, Dense layer, a 50\% drop out, another dense layer, another 50\% drop out, and finally, a fully connected layer with softmax. Although, the network's training accuracy was 93\%, the model validation and test accuracy were less than 50\%. This motivated us to design a new architecture which adapts and improves on the architecture from \cite{daniel}. This neural network architecture possesses 6 convolution layers. Output from each convolution layer is passed through a rectified linear unit (ReLU). The first and the second convolution layers possess 64 filters, the third and the fourth has 128 filters while the fifth and the sixth convolution layer has 256 filters. Each convolution layer has zero padding. We have a max pooling layer after each pair of convolution layer with 10\% dropout for dimensionality reduction and over-fitting reduction respectively. At the end of the six convolution layers are 3 fully connected layers. The last fully connected layer has a softmax activation function which outputs probability distribution for each of the 10 classes.

\subsection{Model Training and Performance}
We used Adam optimizer, weighted categorical cross-entropy loss function, batch size of 20 for each step, and an epoch of 50. Training took about 1 hour, 45 minutes on a NVIDIA Tesla P100 Workstation GPU with 16G RAM. The performance of the model is discussed in Section 6.

\subsection{Image Segmentation through Patch-Based Prediction}
Each satellite image has a dimension of 3200 x 4800 pixels. This can be too large for directly training a CNN network and will therefore require very extensive computational resources and more training time. In addition, there exist no labelled data from Nairobi region for training patches of this image. We therefore pre-processed the dataset by removing cloudy images and performing linear interpolation to fill gaps, and then applied our patch-based prediction technique on each image using our trained model.  The final output is an image, segmented on land-use basis. The visual smoothness of the output image depends on the stride value chosen. We chose a stride value of 32 in this paper (see Figure \ref{seg_map}), as a balance between acceptable visual quality and acceptable computational time.

\section{Post Classification}
In this section, we explain our approach to land-use change computation, image differencing, build-up index (BUI) formulation, and finally forecasting. 

\subsection{Land-Use Square Area Change}
The source of the Google Earth's images for the region we considered is Digital Globe's QuickBird satellite. QuickBird satellite captures the earth's surface at high resolution of 0.65 meters. This information provides a means for translating our pixel values to the actual ground values.
Thus, we simply convert to actual ground values by multiplying by \((0.65 \times 0.65)m^2\).
\[AC = PC  *  (0.65 \times 0.65)m^2\]
where AC is actual change value and PC is pixel change value.

\subsection{Percentage Change}
We compute the percentage change for the segmentation maps using the formulation below:
\[PC = \frac{C_t - C_{t-1}}{ C_{t-1}} \times 100 \]
\(PC\) is percentage change, \(C_t\) is change on the image at time \(t\), and \(C_{t-1}\) is change on the image preceding the first (image at time \(t\)).  
\subsection{Image Differencing}
To determine visual changes between segmented maps, we use image differencing technique. This involves computing the difference between two images by finding the difference between each pixel in each image and generating an image based on the result.

\subsection{Built-Up Index (BUI)}
In order to see the degree of land spaces being used up through the construction of buildings and other infrastructures, we compute BUI for each image. We do this by finding the ratio of the sum of land-use classes with buildings or roads to the sum of the entire land-use classes.
\[BUI = \frac{\sum{CWB}}{\sum{C}}\] 
where CWB is land-use classes with buildings and roads. They include road junction/roundabout, commercial area, dense residential, freeway, parking lot, sparse residential, and medium residential. C is the entire land-use classes.
 
\subsection{Land-Use Forecasting}
We implemented an ARIMA model, we set parameters as follows: lag value of 4 (since we got data on quarterly basis) for autoregression, difference order of 1 to make the time series stationary, and then a moving average value of 0. Parameters were chosen after studying the nature of the series using auto-correlation plots.

\begin{figure}
\vskip 0.2in
\begin{center}
\centerline{\includegraphics[width=8cm, height=5cm]{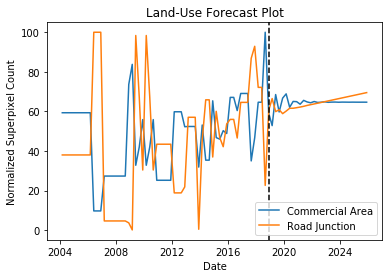}}
\caption{Forecast for Commercial Area and Road Junctions, with evidences of seasonality.}
\label{forecast}
\end{center}
\vskip -0.2in
\end{figure}

ARIMA is a popular and widely used statistical method for time series forecasting. It is capable of capturing different standard temporal structures in time series data.
Due to the limited data points, we fitted the ARIMA model on 60 images and forecast till Dec. 2025; See model performance in Figure \ref{forecast}.
\begin{figure*}
\vskip 0.2in
\begin{center}
\centerline{\includegraphics[width=10cm, height=8cm]{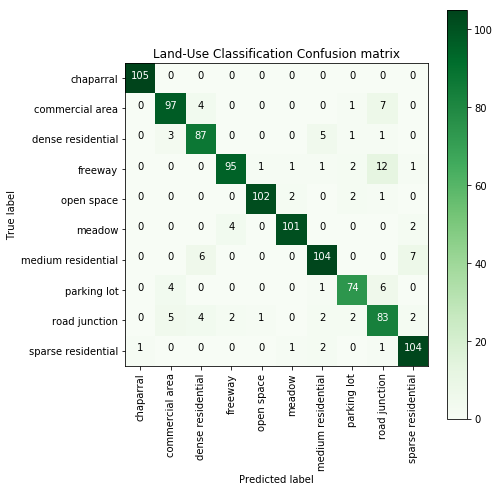}}
\caption{Confusion matrix for model's performance, overall precision  and recall values are 0.907 and 0.908 respectively. }
\label{confusion}
\end{center}
\vskip -0.2in
\end{figure*}

\begin{figure*}
\vskip 0.2in
\begin{center}
\centerline{\includegraphics[width=12.5cm, height=9cm]{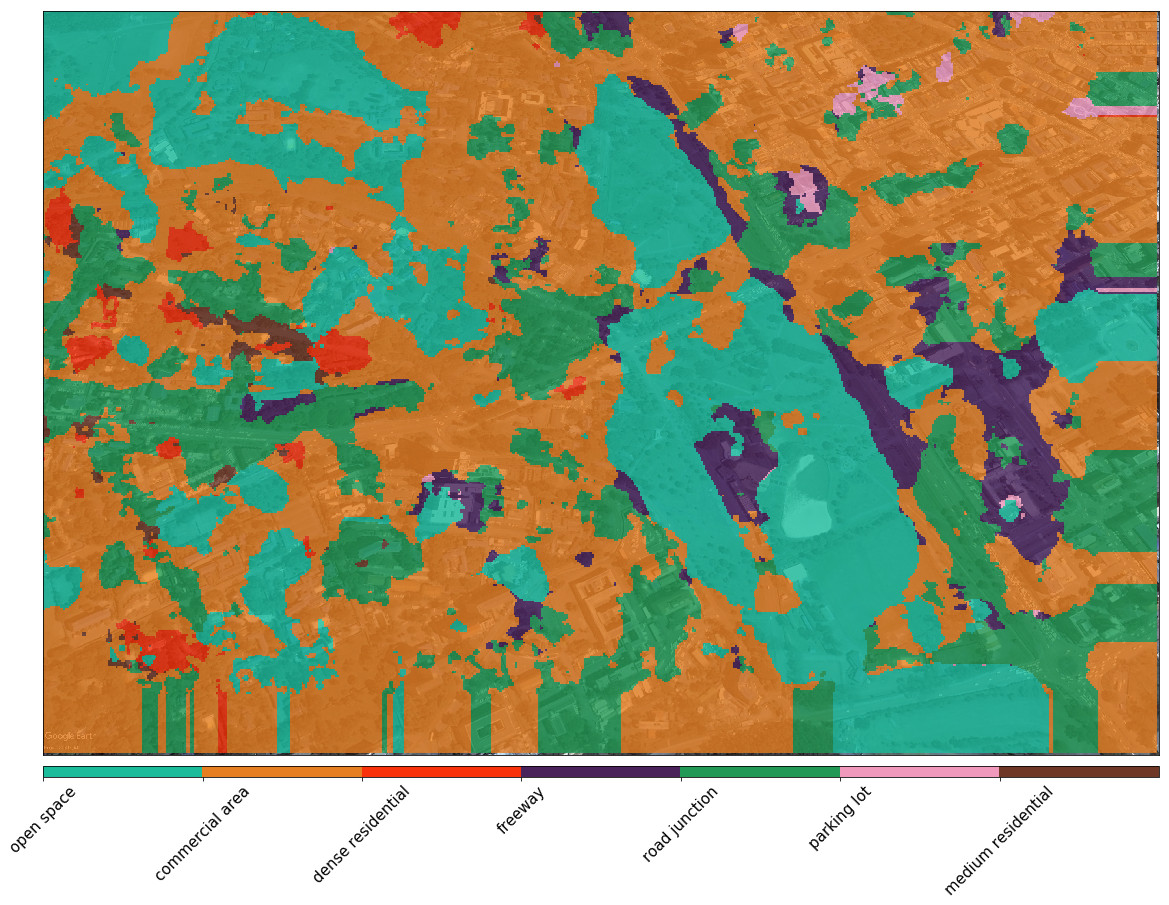}}
\caption{Segmented map from patch-based prediction}
\label{seg_map}
\end{center}
\vskip -0.2in
\end{figure*}
\begin{figure}
\vskip 0.2in
\begin{center}
\centerline{\includegraphics[width=9cm, height=6cm]{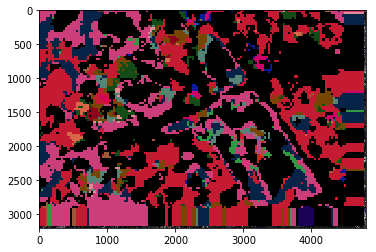}}
\caption{Image difference map between March 2004 and December 2018, the black pixel indicates no change.}
\label{fig:image_diff}
\end{center}
\vskip -0.2in
\end{figure}

\begin{figure}
\vskip 0.2in
\begin{center}
\centerline{\includegraphics[width=8cm, height=5cm]{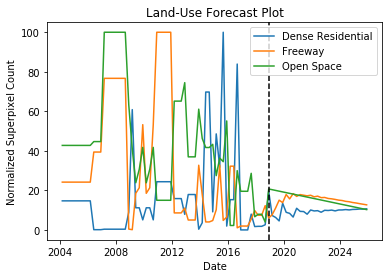}}
\caption{Forecast for few other classes.}
\label{other}
\end{center}
\vskip -0.2in
\end{figure}

Table \ref{tab:table1} and Table \ref{tab:table2} give the change values and forecast values respectively.
\begin{table*}[ht]
\centering
\caption{Land-Use Change from Dec. 2011 to Dec. 2018, BUI of 0.644 and 0.880 for Dec. 2011 and Dec. 2018 respectively}
\label{tab:table1}
\begin{tabular}{ |p{5cm}||p{2cm}|p{2cm}|p{2cm}|p{2cm}|}
 \hline
 \multicolumn{5}{|c|}{Land-Use Change for 2011 to 2018} \\
 \hline
 Land-Use& Dec. 2011 Area \((km^2)\) & Dec. 2018 Area \((km^2)\)& Change in Area \((km^2)\) & Change \% \\
 \hline
 -Chaparrel   & 0.000    &0.000&   0.000&      0.000\\
 -Commercial Area &   3.811&	3.063&	-0.748&	-19.645\\
 -Dense Residential & 0.064&	0.139&	0.075&	116.779\\
 -Freeway    &0.006&	0.088&	0.082&	1356.867\\
 -Open Space (bare land, wetland, plantations)&2.308&	0.755&	-1.552&	-67.254\\
 -Meadow& 0.000  & 0.005   &0.005&  \(\infty\)\\
 -Medium Residential& 0.000  & 0.350 &0.350& \(\infty\)\\
 -Parking lot & 0.073& 0.023&	-0.050&	-68.236\\
 -Road Junctions &0.117&	1.930&	1.812&	1540.102\\
 -Sparse Residential &0.117&	0.025&	-0.091&	-77.938\\
 \hline
\end{tabular}
\end{table*}

\begin{table*}[ht]
\centering
\caption{Land-Use Change Forecast for Dec. 2025, BUI for Dec. 2025 is 0.940}
\label{tab:table2}
\begin{tabular}{ |p{5cm}||p{2cm}|p{2cm}|p{2cm}|p{2cm}|}
 \hline
 \multicolumn{5}{|c|}{Land-Use Change Forecast for Dec. 2025} \\
 \hline
 Land-Use& Dec. 2018 Area \((km^2)\) & Dec. 2025 Area \((km^2)\)& Change in Area \((km^2)\) & Change \% \\
 \hline
 -Chaparrel   & 0.000    &0.000&   0.000&      0.000\\
 -Commercial Area &   3.063&	3.175&	0.112&	3.675\\
 -Dense Residential &0.139&	0.076&	-0.063&	-45.394\\
 -Freeway    &0.088&	0.184&	0.096&	109.600\\
 -Open Space (bare land, wetland, plantations)&   0.755&	0.374&	-0.381&	-50.445\\
 -Meadow& 0.005  &	0.018&	0.012&	221.167\\
 -Medium Residential& 0.350&	0.446&	0.096&	27.604\\
 -Parking lot & 0.023& 0.124& 0.101&	434.633\\
 -Road junction &1.930&	2.131&	0.201&	10.424\\
 -Sparse Residential &0.025&	0.045&	0.019&	76.329 \\
 \hline
\end{tabular}
\end{table*}

\section{Results and Discussion}
We evaluated our model on the NWPU-RESISC45 test dataset which is 15\% of the original dataset and contains 1050 samples. The accuracy of the model on the test dataset was 91.03\% with overall average AUC ROC score of 0.99 and an F1 score of 0.90. This performance is an improvement over the chosen benchmark which employed transfer learning approach using VGGNet-16 and achieved a test accuracy of 90.36\%; See Figure \ref{confusion}. 

The output of our patch-based prediction, which are labelled segmentation maps of land-use, is visually smooth, and provides qualitative evidence of the correctness of the results based on visual inspection; See Figure \ref{seg_map}. Thus, our hypothesis that a model trained on NWPU-RESISC45 dataset would perform well on Google Earth's image patches for regions with similar land-use classes with NWPU-RESISC45 dataset is valid. The visual smoothness of the output images depend on the stride value chosen. For instance, patch-based prediction with stride value of 8 produces better segmentation map in terms of visual smoothness, when compared with segmentation map generated using stride values of 32 or 256.

Figure \ref{fig:image_diff} shows the image differencing output, it highlights regions that have changed and those that have remained unchanged over the years. The black regions indicate regions that did not change, while other colours show that changes occurred. This can help expose rapidly changing environments and stable environments. Figure \ref{forecast} and Figure \ref{other} show the forecast plots for land-use. Visual inspection of these graphs shows evidences of seasonality in some of the land-use classes.

Table \ref{tab:table1} compares land-use in 2011 and 2018. The table shows that commercial area's size reduced, this may be because of the increase in dense residential regions, some commercial buildings must have been modified and now converted to residential houses. It could also be that as the population increased, people built residential houses in the commercial environment. BUI increased from within 2011 to 2018.
Table \ref{tab:table2} shows the predicted values for 2025 December. In this case, commercial area and road junctions/round-abouts are predicted to increase. The predicted BUI for 2025 also indicates an increase. This shows an evidence of infrastructural development.

\section{Conclusion and Future Work}
An efficient machine learning technique for large-scale urban land-use classification has potential to help policy makers and urban planners make decisions. In this paper, we presented a new efficient deep convolutional neural network for urban land-use classification, which applies patch-based prediction to classify satellite images and generate segmentation maps. The NWPU-RESISC45 dataset was used to train the classification network. This network was used to do a patch-based prediction on a dataset that contained 60 satellite images of Kilimani area of Nairobi, acquired from 2004 to 2018 on a quarterly basis. Our model exceeds the performance of the benchmark model by 1\%, but using a significantly simpler CNN which can be trained much more quickly where transfer learning is not suitable. We detected changes in land-use over the period and also computed build-up index (BUI) for each aerial image scene. Forecasting was done using ARIMA model fitted on the 60 data points. Forecast result shows evidences of seasonality. Since developing countries are well represented in our training dataset, we conclude that our model can be applied in any region in sub-Saharan Africa. 

As one of the strengths of this research, our approach saves much time when an optimal value of stride is chosen during patch-based prediction. This is because the we applied the patch-based prediction approach with a simple deep CNN model. This approach is novel for land-use forecasting in sub-Saharan Africa and it can even be extended to some other developing countries in the world.
One of the limitations of this work is that when a very small value of stride (for instance 1) is used during the prediction of a very large aerial image, much time is consumed. Therefore, there is a problem in deciding a stride value that will produce good visual quality in good time. Future works include generating synthetic forecast land-use maps based on historical land-use maps using Generating Adversarial Networks (GAN). We also want to formulate an equation to determine the optimal value of stride for each occasion.
For compatibility and reuse reasons, the shape file format of the land-use classification outputs of this research will be created. This will help researchers and remote sensing engineers easily reuse our outputs in other projects. Finally, we would extend this work to other regions in Africa, and investigate the societal impacts of this land-use forecast in the coming years.

\end{document}